\documentclass[11pt,a4paper]{article}
\usepackage[hyperref]{acl2021}
\usepackage{times}
\usepackage{latexsym}

\usepackage{todonotes}
\usepackage{amsmath}
\usepackage{amssymb}
\usepackage{amsthm}
\newtheorem{exm}{Example}
\usepackage{graphicx}
\usepackage{caption}
\usepackage{multirow}
\usepackage{adjustbox}
\usepackage{subcaption}
\usepackage{booktabs}
\usepackage{array}
\newcolumntype{M}[1]{>{\centering\arraybackslash}m{#1}}
\usepackage{hyperref}
\usepackage{xcolor}

\newcommand{\wtilde}[1]{\stackrel{\sim}{\smash{#1}\rule{0pt}{1.1ex}}}

\usepackage{microtype}

\aclfinalcopy % Uncomment this line for the final submission
\def\aclpaperid{26} %  Enter the acl Paper ID here

\title{Interactive Acquisition of Fine-grained Visual Concepts by Exploiting Semantics of Generic Characterizations in Discourse}

\author{Jonghyuk Park \and Alex Lascarides \and Subramanian Ramamoorthy \\
  School of Informatics, University of Edinburgh \\
  10 Crichton Street, Edinburgh EH8 9AB, UK \\
  \texttt{jay.jh.park@ed.ac.uk}, \texttt{alex@inf.ed.ac.uk}, \texttt{s.ramamoorthy@ed.ac.uk} \\}

\date{}

\begin{document}
\maketitle
\begin{abstract}
Interactive Task Learning (ITL) concerns learning about unforeseen domain concepts via natural interactions with human users.
The learner faces a number of significant constraints: learning should be online, incremental and few-shot, as it is expected to perform tangible belief updates right after novel words denoting unforeseen concepts are introduced.
In this work, we explore a challenging symbol grounding task---discriminating among object classes that look very similar---within the constraints imposed by ITL.
We demonstrate empirically that more data-efficient grounding results from exploiting the truth-conditions of the teacher's generic statements (e.g., ``Xs have attribute Z.'') and their implicatures in context (e.g., as an answer to ``How are Xs and Ys different?'', one infers Y lacks attribute Z).
\end{abstract}

\section{Introduction}
\label{sec:intro}

\begin{figure}
\centering
\hfill
\begin{subfigure}{0.45\textwidth}
    \includegraphics[width=\textwidth]{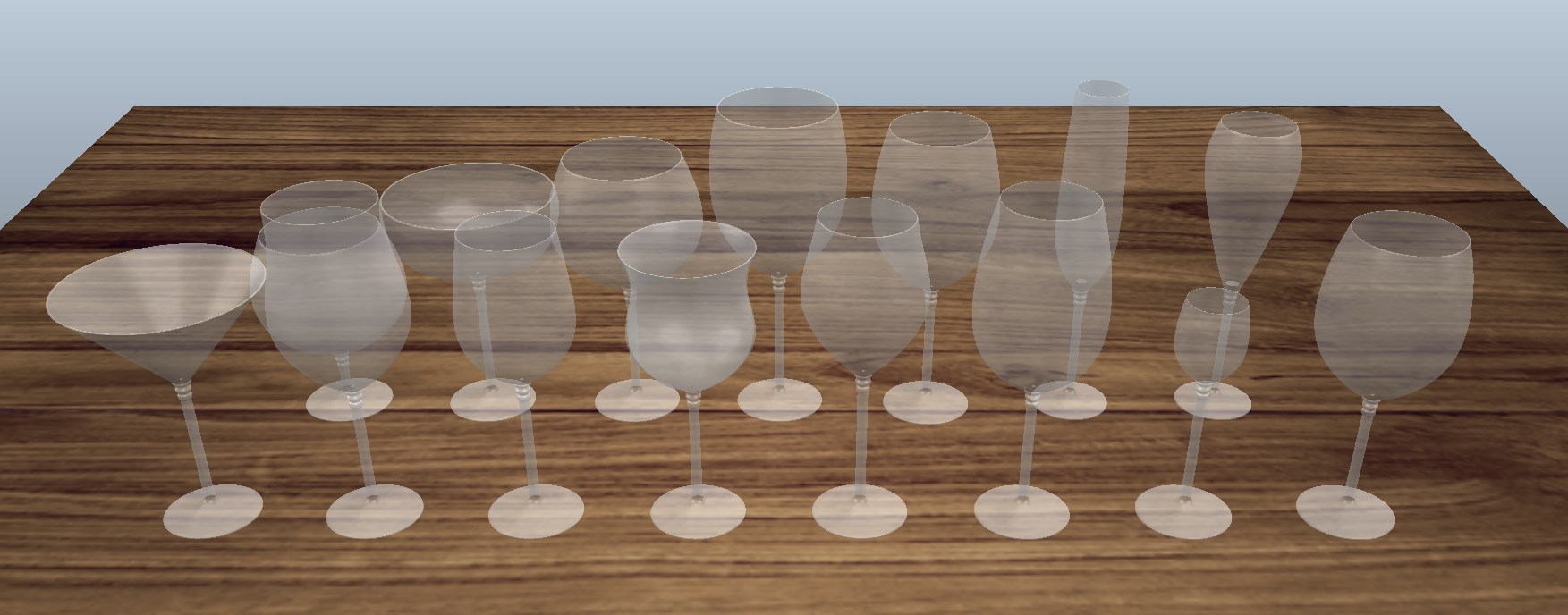}
    \caption{3D models of fine-grained types of glasses.}
    \label{fig:glass_types}
\end{subfigure}
\par\bigskip
\begin{subfigure}{0.5\textwidth}
    \includegraphics[width=\textwidth]{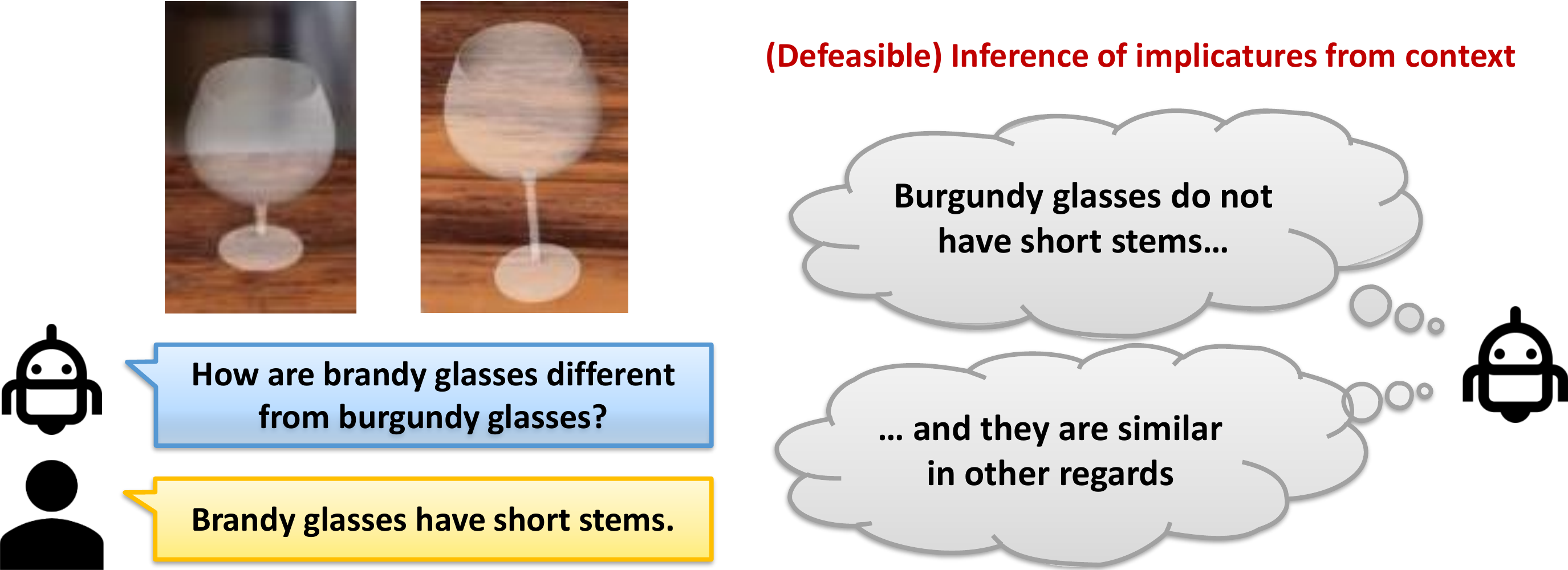}
    \caption{Example interaction between a teacher and a learner discussing generic knowledge about types of glasses.}
    \label{fig:interaction}
\end{subfigure}
\caption{Learning via embodied dialogue in a simulated tabletop domain.}
\end{figure}

Consider a general-purpose robot assistant purchased by a restaurant, which must acquire novel domain knowledge to operate in this particular venue.
For example, the agent must learn to distinguish brandy glasses from burgundy glasses (Fig.~\ref{fig:glass_types}), but these subcategories of glasses are entirely absent from the agent's domain model in its factory setting.
Learning to distinguish among fine-grained visual subcategories is a nontrivial feat \cite{wei2021fine}; most current approaches require careful engineering by ML practitioners, making them unsuitable for lay users to readily inspect and update the robot's domain knowledge.
% Moreover, these methods deploy neural networks that are difficult to interpret, thereby preventing lay users from inspecting the agent’s current domain knowledge with ease.

There are also challenges regarding data efficiency, which using natural language can potentially address \cite{laird2017interactive}.  
A single \textit{generic statement}---e.g., ``Brandy glasses have short stems''---expresses content that would take many visual examples to infer. 
Such statements, given their dialogue context, may also carry additional meaning that is linguistically implicit.
For instance, if the statement ``Brandy glasses have short stems'' is given as an answer to the contrastive question ``How are brandy glasses and burgundy glasses different?'', then it implies that burgundy glasses don't have short stems, and also, defeasibly, that these two types of glasses are similar in other conceivable respects \cite{grice1975, asher2013implicatures}.
Vision processing models that exploit natural language data exist \cite{he2017fine, xu2018fine, chen2018knowledge, song2020bi}, but they generally treat language as supplementary signals for augmenting training examples, rather than leveraging a range of symbolic inferences licensed by purposeful utterances in dialogue.

In this work, we develop an \textit{interactive} symbol grounding framework, in which the teacher presents to the learner evidence for grounding during embodied dialogues like those illustrated in Fig.~\ref{fig:interaction}.
The framework is based on a highly modular neurosymbolic architecture, in which subsymbolic perceptual inputs and symbolic conceptual knowledge obtained during dialogues gracefully combine. 
% combine by means of logical reasoning.
We run proof-of-concept experiments to show that agents that exploit semantic and pragmatic inferences from generic statements in discourse outperform baselines that don't exploit semantics and pragmatics, or don't exploit symbolic inference at all.

\section{Related Work}

In fine-grained image analysis (FGIA), a model learns to distinguish (patches of) images of subcategories that belong to the same basic category.
FGIA is challenging because images exhibit small inter-class variance and large intra-class variance, and
labeling often requires specific domain expertise, hence high annotation costs \cite{wei2021fine}.

A natural approach to FGIA is to utilize information of different modalities, including unstructured text descriptions \cite{he2017fine, song2020bi}, structured knowledge bases \cite{xu2018fine, chen2018knowledge} and human-edited attention maps \cite{duan2012discovering, mitsuhara2021embeddingHK}.
However, to our knowledge, no existing FGIA models exploit NL generic statements provided \textit{in vivo} during natural dialogues.
Existing interactive FGIA methods \cite{branson2010visual, wah2011multiclass, wah2014similarity, cui2016fine} query humans to refine predictions from off-the-shelf vision models at inference time but do not update the grounding models.
In contrast, our framework supports continuous learning, updating the grounding model as and when the teacher says something noteworthy.

Our use case, described in \S\ref{sec:intro}, can be subsumed under the framework of Interactive Task Learning (ITL; \citealp{laird2017interactive}).
Motivated by scenarios where unforeseen changes may happen to the domain after deployment, the core goal of ITL is to acquire novel concepts that the learner is unaware of but are critical to task success.
ITL systems gather evidence from \textit{natural embodied interactions} with a teacher that take place while the learner tries to solve its task.
Thus a key desideratum in ITL is that learning should be online and incremental: the learner should change its beliefs and behaviours whenever the teacher provides guidance.

Natural language is a common mode of teacher-learner interaction in ITL \cite{kirk2016learning,she2017interactive}.
Accordingly, several ITL works draw inspiration from linguistic theories to make learning more effective and efficient.
While the formal semantics of quantifiers and negation \cite{rubavicius2022interactive} and of discourse coherence \cite{appelgren2020interactive} has been explored in ITL settings, none of the works in the ITL literature have investigated the utility of exploiting the logical inferences licenced by the semantics and pragmatics of contrastive questions and their generic statement answers.

\section{Agent architecture}
\label{sec:arch}

\begin{figure}
\centering
\includegraphics[width=0.45\textwidth]{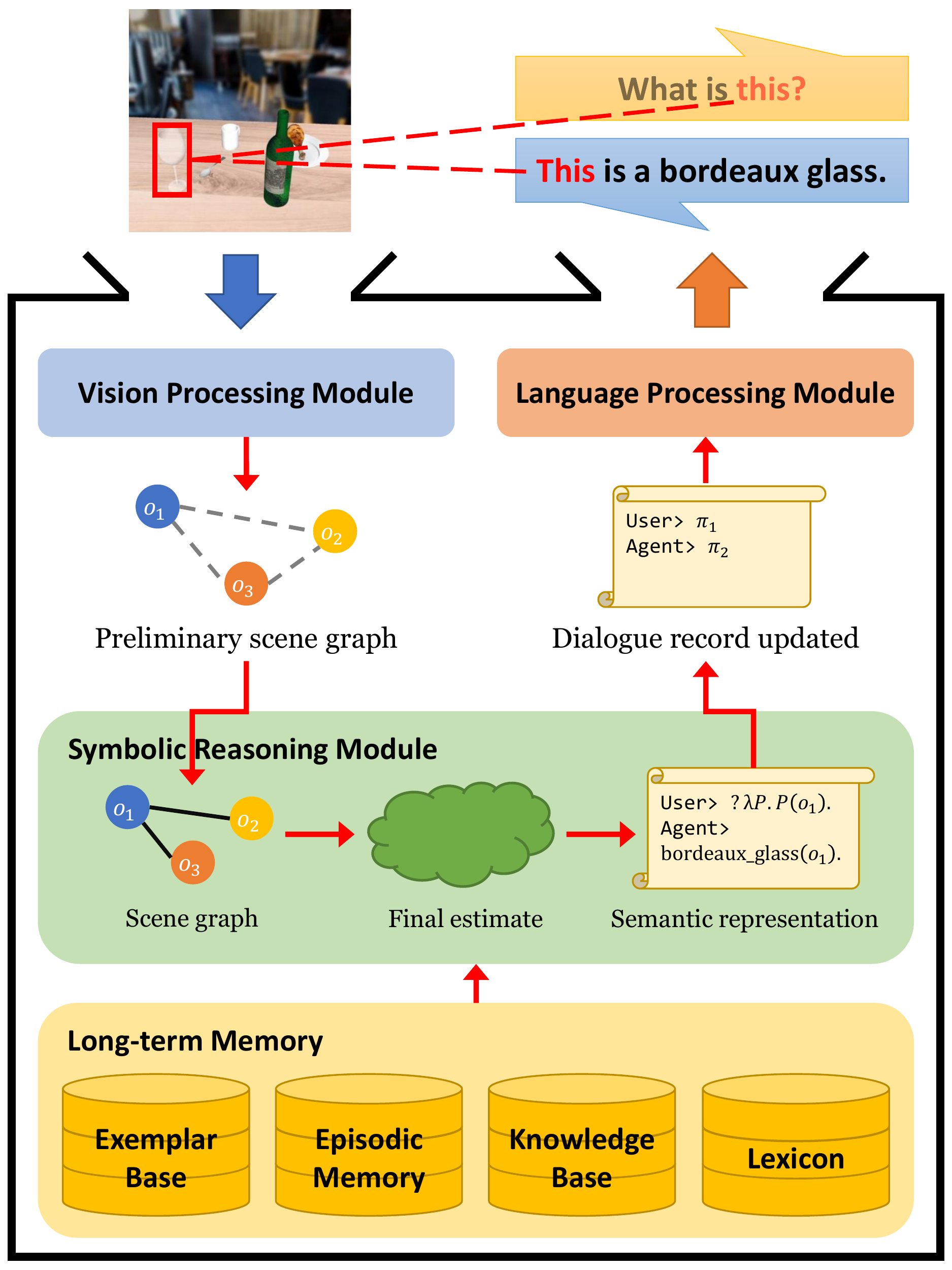}
\caption{Overview of the architecture in inference mode, in which the component modules interact to generate an answer to a user question.}
\label{fig:arch}
\end{figure}

Fig.~\ref{fig:arch} illustrates our neurosymbolic architecture for situated ITL agents that can engage in extended dialogues with a teacher.
Its design enables both subsymbolic-level learning of visual concepts from perceptual inputs (``This looks like a X'') and symbolic-level learning and exploitation of relational knowledge between concepts (``Xs generally have attribute Z'') \textit{during} task execution.
Here we stress that our proposed approach is not in direct competition with wide-coverage neural vision-language models, but actually complements them.  
% However large, such models aren't guaranteed to cover all the required concepts in arbitrary domains and for arbitrary tasks.  
As an ITL framework, we offer a coping mechanism, to be employed when an existing pre-trained model is deployed in a domain where concepts are frequently introduced and changed, requiring the model to quickly adapt with only a few exemplars of unforeseen concepts. 

\subsection{Vision processing module}
\label{sec:arch:vis}

\begin{figure}
\centering
\includegraphics[width=0.48\textwidth]{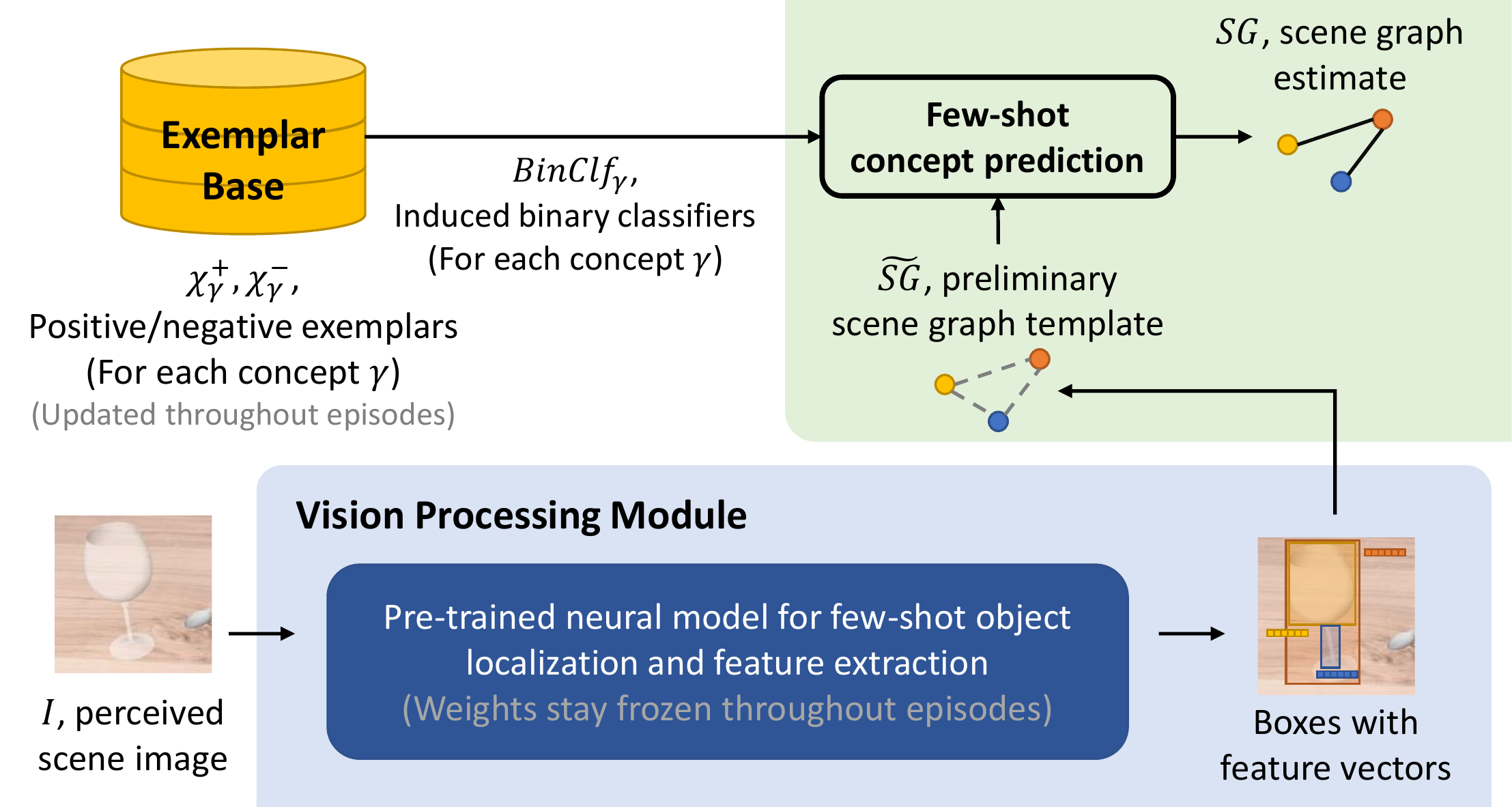}
\caption{Abridged illustration of the few-shot scene graph generation process (Full version in Appendix~\ref{app:vision})}
\label{fig:vis_simp}
\end{figure}

Given a visual scene perceived by the vision sensor, the agent first summarizes the raw input into a graph-like data structure (\textit{scene graph} hereafter).
A scene graph $SG$ encodes a set of salient objects in the scene with their distinguishing features and their pairwise relationships, serving as the agent's internal, abstracted representation of the scene.

Our architecture makes exemplar-based few-shot predictions to generate scene graphs, so as to quickly learn novel visual concepts after a few training instances in an online, incremental fashion.
Specifically, our vision processing module employs a neural model extended from Deformable DETR \cite{zhu2021deformable}, trained to learn distinct low-dimensional metric spaces for each concept type (object class/attribute/relation).
The module makes binary concept predictions based on similarity distances between embedded vectors.
As illustrated in Fig.~\ref{fig:vis_simp}, the role of the vision module is to process an RGB image input $\mathcal{I}$ into a preliminary scene graph template $\wtilde{SG}$.
The template is further processed along with the agent's store of concept exemplars in its long-term memory (\S\ref{sec:arch:mem}) to yield $SG$.
For further details about the inner working of the neural vision module and the translation process from $\wtilde{SG}$ to $SG$, refer to Appendix~\ref{app:vision}.

\subsection{Language processing module}
The language processing module parses natural language utterances into formal semantic representations, maintains dialogue records, and generates natural language utterances as needed.
For controlled experiments, we constrain our attention to a class of simple sentences that discuss primarily two types of information: 1) instance-level descriptions about conceptual identities of scene objects (e.g., ``This is a brandy glass'', ``This has a wide bowl''); and 2) relational knowledge about generic properties shared across instances of the same concepts (e.g., ``Brandy glasses have short stems'').

More formally, we represent the propositions expressed by NL sentences via a simple antecedent-consequent pair (\textsc{prop} hereafter).
\textsc{prop}s draw on a first-order language $\mathcal{L}$ which includes constants referring to objects in the visual scene and predicate symbols for their classes, attributes and pairwise relations (i.e., visual concepts from \S\ref{sec:arch:vis}).
Indicative NL sentences are generally represented with a \textsc{prop} $\psi=Ante\Rightarrow Cons$, where $Ante$ and $Cons$ are each a $\mathcal{L}$-formula (for $\psi$, we refer to these as $Ante(\psi)$ and $Cons(\psi)$).
$Ante(\psi)$ is empty (and thus omitted) if $\psi$ represents a non-conditional, factual statement.
Further, we notate a \textsc{prop} that stands for a generic characterization with a `generic quantifier' $\mathbb{G}$.
For example, the sentences ``$o$ is a brandy glass'' and ``Brandy glasses have short stems'' are translated into \textsc{prop}s respectively as $brandyGlass(o)$ and $\mathbb{G} O.brandyGlass(O)\Rightarrow haveShortStem(O)$.\footnote{In the interest of brevity and simplicity, we have translated ``have short stems'' into an `agglomerate' predicate $haveShortStem$ in this text. This is contrary to the actual implementation, where we introduced the concepts $stem$, $short$ (unary predicates) and $have$ (binary predicate) as elementary units. See Appendix~\ref{app:fol_props} for a more accurate exposition.}

We represent questions (\textsc{ques} hereafter) following notation similar to \citet{groenendijk:stokhof:1982}.
The answer to a polar question, represented as $?\psi$, is $\psi$ (if true) or $\neg \psi$ (if false).
Answers to a \textit{wh}-question $?\lambda X.\psi(X)$ provide values $a$ of $X$ that make $\psi[X/a]$ true (i.e., all occurrences of $X$ in $\psi$ are substituted with $a$).
% A notable exception would be the treatment of questions in the form of ``How are Xs and Ys different?'', whose true answers must identify properties of first-order predicates.
For the question ``How are $p_1$ and $p_2$ different?'', we avoid the complexity of higher-order formal languages and simply introduce a reserved formalism $?\text{\texttt{conceptDiff}}(p_1,p_2)$, which our implemented dialogue participants can handle by invoking a dedicated proecdure.
The answer to $?\text{\texttt{conceptDiff}}(p_1,p_2)$ is the set of attributes that all objects of class $p_1$ have and $p_2$ lack, and \textit{vice versa}.

The language processing module is implemented as a pipeline with two components: an off-the-shelf large-coverage parser of the English Resource Grammar \cite{copestake2000open} followed by manual heuristics that map the parser's outputs to the above forms, as required by the symbolic reasoner (see \S\ref{sec:arch:symb}). 
The module also keeps track of the current dialogue history as a sequence of utterances: each one logged as a \textsc{prop} or \textsc{ques}, its NL surface form and its speaker.

\subsection{Long-term memory module}
\label{sec:arch:mem}

Our agent stores new knowledge acquired over the course of its operation in its long-term memory.
We implement four types of knowledge storage: visual exemplar base (XB), symbolic knowledge base (KB), episodic memory and lexicon.

\paragraph{Visual XB}
For each visual concept $\gamma$, the visual XB stores $\chi_\gamma^+$ and $\chi_\gamma^-$, a set of positive/negative exemplars worth remembering.
The exemplars serve as the basis of the agent's few-shot prediction capability as mentioned in \S\ref{sec:arch:vis}.
The visual XB is expanded each time the agent makes an incorrect prediction.
Specifically, when the learner incorrectly states ``This is $\tilde{\gamma}$'', the teacher provides a corrective response, saying ``This is not $\tilde{\gamma}$, this is $\gamma$'', thereby augmenting $\chi_\gamma^+$ and $\chi_{\tilde{\gamma}}^-$.
New sets $\chi^{+/-}_\gamma$ are created whenever the teacher introduces a novel concept $\gamma$ via a neologism.

\paragraph{Symbolic KB}
The symbolic KB is a collection of generic \textsc{prop}s describing relations between symbolic concepts, such as $\mathbb{G} O.brandyGlass(O)\Rightarrow haveShortStem(O)$.
Each KB entry is annotated with the source of the knowledge: a generic rule may be explicitly uttered by the teacher or inferred as an implicature, given the dialogue context.
We'll discuss how the learner can extract unstated knowledge in \S\ref{sec:task:dial:lstrat} in further detail.

\paragraph{Episodic memory}
The episodic memory stores the summary of each episode of situated interactions between the agent and the teacher.
%% The primary utility of the stored episodes in our approach is that they serve as `test cases', which can validate generic rules that are inferred as implicatures and thus may be potentially incorrect (\S\ref{sec:task:dial:lstrat}).

\paragraph{Lexicon}
The lexicon stores a set of content words the teacher introduces into the discourse, along with linguistic metadata like part-of-speech.
% We assume simple one-to-one correspondence between symbols and their denotations.

\subsection{Symbolic reasoning}
\label{sec:arch:symb}

For symbolic reasoning, we employ a probabilistic variant of a logic programming\footnote{In contrast to first-order logic, logic programming is based on the notion of \textit{minimal models}, where any true atom must be justified (founded) by a clause in the logic program.} technique known as answer set programming (ASP; \citealp{lifschitz2008what}).
The formalism of ASP represents a reasoning problem as a \textit{normal logic program} that consists of rules of the following form:
\begin{equation}
a\leftarrow b_1,\ldots,b_m,\text{\texttt{not} }c_1,\ldots,\text{\texttt{not} }c_n.  \label{eq:rule}
\end{equation}
where the rule head atom $a$ and the rule body atoms $\{b_i\}_{i=1}^m,\{c_j\}_{i=1}^n$ can be propositional or (quantifier-free) first-order logic formulas.
An intuitive reading of the rule, by itself, is that $a$ is logically justified if and only if all of the positive body atoms $\{b_i\}_{i=1}^m$ hold and none of the negative body atoms $\{c_j\}_{i=1}^n$ are proven to hold.
For instance, the ASP rule $fly(X)\leftarrow bird(X),\text{\texttt{not} }abnormal(X)$ would roughly correspond to the meaning of the generic NL statement ``Birds (generally) fly''.
A rule whose head is empty ($\bot$) represents an \textit{integrity constraint} that its rule body should not hold in answer models.

In probabilistic ASP \cite{lee2016weighted}, each rule is associated with a weight, such that possible worlds satisfying a set of rules with higher total weights are assigned greater probability.
Thus a rule may be violated at the expense of its weight.
Formally, a probabilistic ASP program $\Pi=\{w:R\}$ is a finite set of weighted rules where $R$ is a rule of the form (\ref{eq:rule}) and $w$ is its associated weight value.
The probability of a possible world $I$ according to $\Pi$ is computed via a log-linear model on the total weight of rules in $\Pi_I$, where $\Pi_I$ is the maximal subset of $\Pi$ satisfiable by $I$. %\citet{lee2016weighted}.
\begin{align}
\label{eq:asp1}
W_\Pi(I)&=exp\Biggl(\sum_{w:R\in\Pi_I} w\Biggl) \\
\label{eq:asp2}
P_\Pi(I)&=\frac{W_\Pi(I)}{\sum_{J\in \text{possible worlds by }\Pi} W_\Pi(J)}
\end{align}
For more rigorous technical definition, refer to \citet{lee2016weighted}.

Each symbol grounding problem is cast into an appropriate program as follows.
First, serialize the learner's visual observations contained in the scene graph $SG$ into $\Pi_O=\{\text{logit}(s):\gamma(o_1,...).\}$, where each $\gamma(o_1,...)$ is a visual observation in $SG$ with confidence score $s\in[0,1]$.
Then we export the KB into a program $\Pi_K$, built as follows:
\begin{itemize}
    \item For each KB entry $\kappa$, add to $\Pi_K$: 
    \begin{multline*}
    \text{logit}(U_d):\bot\leftarrow Ante(\kappa),\text{\texttt{not} }Cons(\kappa).
    \end{multline*}
    which penalizes `deductive violation' of $\kappa$.
    \item For each set of KB entries $\{\kappa_i\}$ that share identical $Cons(\kappa_i)$, add to $\Pi_K$:
    \begin{multline*}
        \text{logit}(U_a):\bot\leftarrow Cons(\kappa_i),\\ \bigwedge_{\kappa_i}\{\text{\texttt{not} }Ante(\kappa_i)\}
    \end{multline*}
    which penalizes failure to explain $Cons(\kappa_i)$.
\end{itemize}
Here, $U_d,U_a\in[0,1]$ are parameters encoding the extent to which the agent relies on its symbolic knowledge; we use $U_d=U_a=0.95$ in our experiments.
For instance, the KB consisting of a single \textsc{prop} parsed from ``Brandy glasses have short stems'' will be translated into $\Pi_K$ consisting of the two rules (\ref{rule:deduc}) and (\ref{rule:abduc}) in Example~\ref{exm:1} below.
Finally, the program $\Pi=\Pi_O\cup\Pi_K$ is solved using a belief propagation algorithm \cite{shenoy1997binary} modified to accommodate the semantics of logic programs.

\begin{exm}
\label{exm:1}
The program $\Pi$ below encodes a scenario where the agent sees an object $o_1$ and initially estimates $o_1$ is equally likely to be a brandy or burgundy glass. The agent also notices with high confidence it has a short stem, and knows brandy glasses have short stems:
\begin{align}
\text{\emph{logit}}(0.61) &:\enspace brandyGlass(o_1). \\
\text{\emph{logit}}(0.62) &:\enspace burgundyGlass(o_1). \\
\text{\emph{logit}}(0.90) &:\enspace haveShortStem(o_1). \\
\text{\emph{logit}}(0.95) &:\enspace \bot\leftarrow brandyGlass(O), \notag\\
    &\qquad\qquad\text{\texttt{not} haveShortStem(O)}. \label{rule:deduc}\\
\text{\emph{logit}}(0.95) &:\enspace \bot\leftarrow haveShortStem(O), \notag\\
    &\qquad\qquad\text{\texttt{not} brandyGlass(O)}. \label{rule:abduc}
\end{align}
This results in $P_\Pi(brandyGlass(o_1))=0.91$, whereas $P_\Pi(burgundyGlass(o_1))=0.62$. Thus the agent forms a stronger belief that $o_1$ is a brandy glass than it is a burgundy glass.
\end{exm}

See Appendix~\ref{app:more_asp_exs} for more examples.

\section{Interactive Visual Concept Acquisition}

\subsection{Task description}

In our symbol grounding task, each input is a tuple $x_i = (\mathcal{I}_i, b_i)$, where $\mathcal{I}_i\in [0,1]^{3\times H\times W}$ is an RGB image and $b_i$ is a specification of a bounding box encasing an object in $\mathcal{I}_i$.
That is, $x_i$ is essentially reference to an object in an image.
The task output $y_i$ is dependent on two possible modes of querying the agent about the identity of the object referenced by $x_i$.
The first `polar' mode amounts to testing the agent’s knowledge of a concept in isolation (i.e., ``Is this a X?''; so $y_i$ is yes or no).
The second `multiple-choice' mode demands the agent selects a single object class $y_i$ describing the object among possible candidates (i.e., ``What is this?'', and $y_i$ is a class).
The teacher's response to $y_i$ is dependent on the content of $y_i$ and the teacher's dialogue strategy as described in \S\ref{sec:task:dial:tstrat}.
The learner updates its symbol grounding model from the teacher's moves using the methods described in \S\ref{sec:arch} and \S\ref{sec:task:dial:lstrat}.  

As mentioned earlier, the agent's domain model may entirely lack the concept of interest for labelling $x_i$.
The agent acquires unforeseen concepts via teacher utterances.
For example,  if ``This ($x_i$) is a brandy glass''  introduces the agent to the unforeseen concept ``brandy glass'',  then $\mathit{BrandyGlass}$ is added to $\mathcal{L}$ and the visual XB is augmented with newly generated sets $\chi_{\mathit{BrandyGlass}}^+=\{x_i\}$ and $\chi_{\mathit{BrandyGlass}}^-=\varnothing$.

\subsection{Flow of dialogues}
\label{sec:task:dial}

\begin{figure}
\centering
\includegraphics[width=0.45\textwidth]{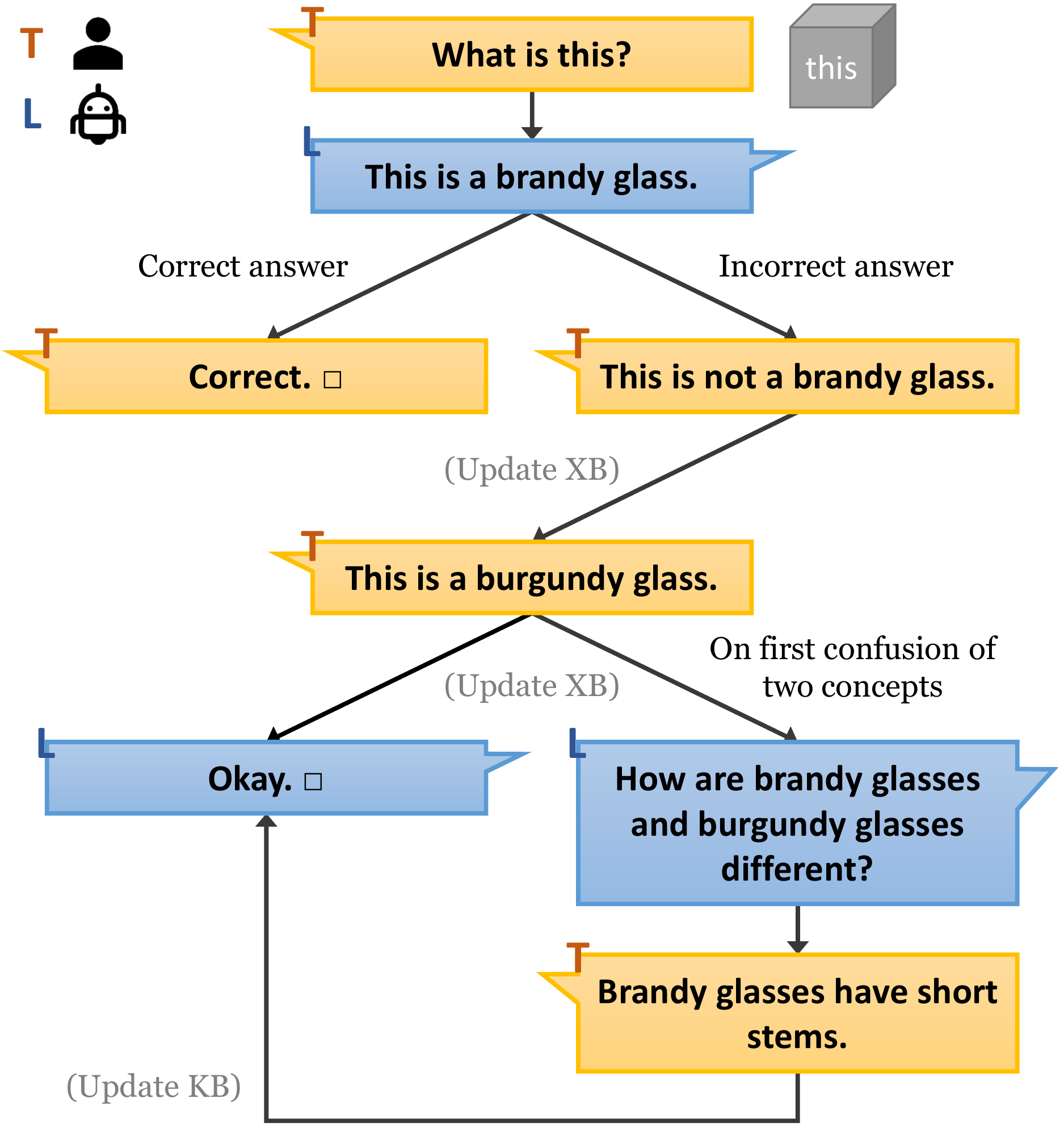}
\caption{Flowchart covering the range of training dialogues modeled in this study. $\square$ signals termination of an interaction episode.}
\label{fig:dialogues}
\end{figure}

We focus on a family of dialogues illustrated in Fig.~\ref{fig:dialogues}.
As depicted, each interaction episode is initiated by a teacher query.
Dialogues will proceed according to the learner's responses and the teacher's strategy. 
In this research, we want to investigate how different interaction and learning strategies affect learning efficiency.  

\subsubsection{Teacher's strategy options}
\label{sec:task:dial:tstrat}

The teacher starts off each interaction episode by presenting an instance $o$ of some visual concept $p$, querying the learner with a probing \textsc{ques} ``$?\lambda P.P(o)$''.\footnote{Note that the expression $?\lambda P.P(o)$ does not fully capture the intended meaning of ``What is this?'' in its own right, since the discourse contexts set up additional semantic/pragmatic constraints on what counts as acceptable answers. We have approximated those constraints via our pre-defined dialogue strategies}
If the learner provides the correct answer as the \textsc{prop} ``$p(o)$'', the teacher responds ``correct'' and the episode terminates without agent belief updates.
Otherwise, if the learner provides an incorrect answer, ``$\tilde{p}(o)$'' or ``I am not sure'', the teacher needs to provide some corrective information so that the learner can adjust its beliefs.
We implement and compare the following variations in the teacher's response, in increasing order of information content:
\begin{itemize}
    \item \textsf{minHelp}: Provides only boolean feedback to the learner's answer, i.e., ``$\neg\tilde{p}(o)$''.
    \item \textsf{medHelp}: In addition to \textsf{minHelp}, provides the correct answer label, saying ``$p(o)$''.
    \item \textsf{maxHelp}: In addition to \textsf{medHelp}, provides a set of generic \textsc{prop}s that characterize $p$ or $\tilde{p}$. The feedback is provided after the learner's \textsc{ques} ``$?\text{\texttt{conceptDiff}}(p,\tilde{p})$'', asked once on the first confusion between $p$ and $\tilde{p}$.
\end{itemize}
The generic \textsc{prop}s provided by \textsf{maxHelp} teachers originate from the teacher's domain knowledge, which we assume here to be correct and exhaustive.
The set of \textsc{prop}s to be delivered is computed as the symmetric difference between the set of properties of $p$ versus that of $\tilde{p}$ (see Appendix~\ref{app:glasses} for an example).
\textsf{minHelp} and \textsf{medHelp} serve as vision-only baselines since only concept exemplars with binary labels are communicated as teaching signal.

\subsubsection{Learner's strategy options}
\label{sec:task:dial:lstrat}

Another dimension of variation we model is the learner’s strategy for interpreting generic statements within dialogue contexts.
Note that the variation in this dimension is meaningful only when the teacher deploys the \textsf{maxHelp} strategy, thereby allowing exploitation of generic statements.

In human dialogues, interlocutors infer, and speakers exploit, \textit{implicatures} that are validated by linguistically explicit moves, given the context of utterance \cite{grice1975}. 
As a core contribution of this study, we model how generic statements given as an answer to a question about similarities and differences give rise to certain implicatures \cite{asher2013implicatures} that can be exploited for more data-efficient learning.

\begin{table}
\small
\centering
\begin{adjustbox}{width=0.48\textwidth}
\begin{tabular}{l|l}
    \toprule
    \multicolumn{2}{c}{Situation} \\
    \midrule
    Confusion & \texttt{brandy glass} vs. \texttt{burgundy glass} \\
    \midrule
    Teacher input & ``Brandy glasses have short stems.'' \\
    \midrule
    Current KB & $\mathbb{G} O.brandyGlass(O)\Rightarrow haveWideBowl(O)$ \\
    \bottomrule
    \multicolumn{1}{c}{} & \\
    \multicolumn{1}{c}{} & \\
    \toprule
    \multicolumn{1}{c|}{Strategy} & \multicolumn{1}{c}{New KB entries added} \\
    \midrule
    \multirow{1}{*}{\textsf{semOnly}} & $\mathbb{G} O.brandyGlass(O)\Rightarrow haveShortStem(O)$ \\
    \midrule
    \multirow{2}{*}{\textsf{semNeg}} & $\mathbb{G} O.brandyGlass(O)\Rightarrow haveShortStem(O)$ \\
    & $\color{red} \mathbb{G} O.burgundyGlass(O)\Rightarrow\neg haveShortStem(O)$ \\
    \midrule
    \multirow{3}{*}{\textsf{semNegScal}} & $\mathbb{G} O.brandyGlass(O)\Rightarrow haveShortStem(O)$ \\
    & $\color{red} \mathbb{G} O.burgundyGlass(O)\Rightarrow\neg haveShortStem(O)$ \\
    & $\color{blue} \mathbb{G} O.burgundyGlass(O)\Rightarrow haveWideBowl(O)$ \\
    \bottomrule
\end{tabular}
\end{adjustbox}
\caption{An example of how different learner strategies update their KBs from the teacher's generic statement feedback after the learner has confused a burgundy glass for a brandy glass. The learner has already learned that burgundy glasses have wide bowls. \textsc{prop}s in black is obtained from the teacher's NL utterance; \textsc{prop}s in \textcolor{red}{red} from `negative' implicatures ($\psi^{neg}$ from $\psi$) as demanded by coherence; and \textsc{prop}s in \textcolor{blue}{blue} from scalar implicatures ($\kappa^{scl}$ from $\kappa$).}
\label{tab:impl_ex}
\end{table}

Suppose a question ``How are X and Y different?'' is answered with a generic statement ``Xs have attribute Z''.
The following implicatures can arise from this discourse context: 1) ``Ys do not have attribute Z'', and 2) ``X and Y are otherwise similar''.
The former follows from the assumption that the generic is a coherent answer to a contrastive question \cite{asher2003logics}. 
The latter, which arguably is more defeasible \cite{grice1975}, is what's known as a scalar implicature: if there were other important differences that the learner should know, then Gricean maxims of conversation predict that the teacher would have included them in the answer as well.

For a \textsc{prop} $\psi$, let $\psi^{p\leftrightarrow q}$ denote a \textsc{prop} which is identical to $\psi$ except that occurrences of the predicate $p$ are all substituted with the predicate $q$ and \textit{vice versa}.
We consider the following strategies the learner can take when interpreting a set of generic \textsc{prop}s $\{\psi_i\}$ provided during an episode:
\begin{itemize}
    \item \textsf{semOnly}: Simply add all $\psi_i$'s to KB.
    \item \textsf{semNeg}: In addition to \textsf{semOnly}, infer a generic \textsc{prop} $\psi_i^{neg}=Ante(\psi_i^{p\leftrightarrow\tilde{p}})\Rightarrow\neg Cons(\psi_i^{p\leftrightarrow\tilde{p}})$ for each $\psi_i$ given as answer to $?\text{\texttt{conceptDiff}}(p,\tilde{p})$. 
    \item \textsf{semNegScal}: In addition to \textsf{semNeg}, infer a generic \textsc{prop} $\kappa^{scl}=Ante(\kappa^{p\leftrightarrow\tilde{p}})\Rightarrow Cons(\kappa^{p\leftrightarrow\tilde{p}})$ for each KB entry $\kappa$ that has either $p$ or $\tilde{p}$ mentioned, only if $\kappa^{scl}$ is not inconsistent with any of $\psi_i$'s or $\psi_i^{neg}$'s.
\end{itemize}
For example, consider the example situation illustrated in Tab.~\ref{tab:impl_ex}.
The \textsf{semNeg} learner adds ``Burgundy glasses do not have short stems'' to its KB, and \textsf{semNegScal} in addition adds ``Brandy glasses have X'' for every property X that, according to its KB, burgundy glasses have (e.g., wide bowls).

While the \textsf{semNeg} inference stems from the demand that the teacher's move is a coherent answer \cite{asher2003logics}, the scalar implicatures inferred by \textsf{semNegScal} are defeasible presumptions \cite{grice1975}.
That is, \textsf{semNegScal} risks misunderstanding the teacher's intended meaning, inferring general rules that are incorrect---yet cancellable (see Appendix~\ref{app:impl_fail} for an example failure case).
Subsequent pieces of refuting evidence may falsify the inferred implicatures without rendering the conversation incoherent.
Therefore, we equip our agents with some risk management faculty that can assess and reject contents of scalar implicatures.
This is achieved by periodically testing KB entries whose origin is solely from scalar implicatures, rejecting those whose counterexamples can be found in the episodic memory.

\section{Experiments}

\subsection{Evaluation Scheme}

We run a suite of experiments that evaluate the data efficiency of the learner’s and teacher’s strategies from \S\ref{sec:task:dial}.
Results are averaged over multiple sequences of interaction episodes for each of five combinations of teacher's and learner's strategies: \textsf{minHelp}, \textsf{medHelp}, \textsf{maxHelp+semOnly}, \textsf{maxHelp+semNeg} and \textsf{maxHelp+semNegScal}.
Each episode-initial probing question ``$?\lambda P.P(o)$'' is associated with a randomly selected instance $o$ of a concept selected from a round-robin of the target concepts to be acquired.
For controlled random selections of concept instances and shuffling of the round-robin, 40 seeds are shared across different configurations.
Each sequence continues until the learner makes $N_t$ mistakes in total.

As is common in ITL scenarios, training and inference are fully integrated.
Learning has to take place during use whenever the teacher imparts information.
In this work, we evaluate our learners by having them take `mid-term exams' on a separate test set after every $N_m$ mistakes made ($N_m\leq N_t$).
The mid-term exams comprise binary prediction problems ``$?p(o)$'' asked per every target concept $p$ for each test example $o$, and we collect confidence scores between 0 and 1 as response.
The primary evaluation metric reported is mean average precision (mAP)\footnote{Mean of areas under interpolated precision-recall curves.}; we do not use an F1 score because we are more interested in relative rankings between similar-looking concepts than the learners' absolute performances at some fixed confidence threshold.
We also report averaged confusion matrices collected for the sequence-final exams (partially in Fig.~\ref{fig:confMats}, fully in the supplementary material).

\begin{figure}
\centering
\begin{subfigure}{0.45\textwidth}
    \includegraphics[width=\textwidth]{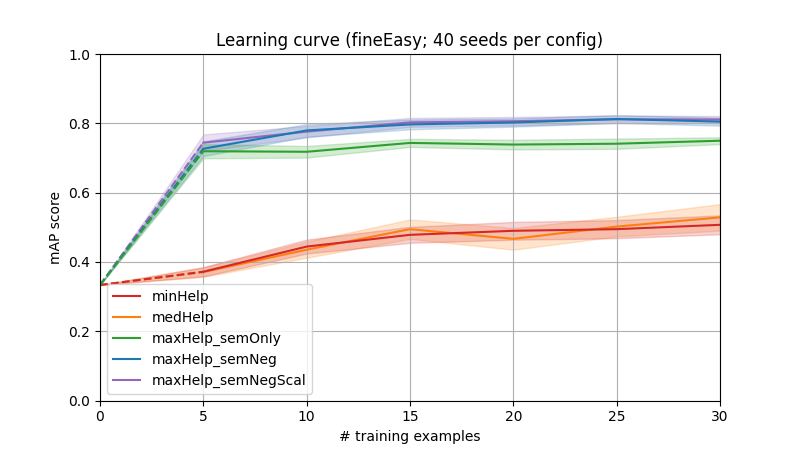}
    \caption{\textsf{fineEasy} difficulty (three glass types)}
\end{subfigure}
\par\bigskip
\begin{subfigure}{0.45\textwidth}
    \includegraphics[width=\textwidth]{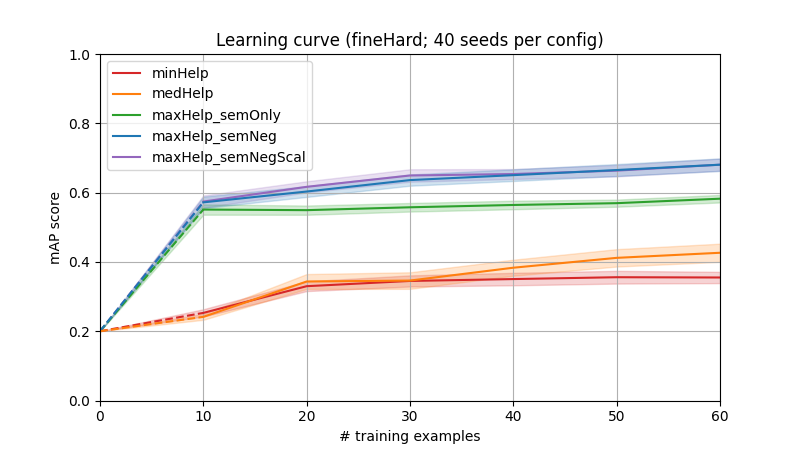}
    \caption{\textsf{fineHard} difficulty (five glass types)}
\end{subfigure}
\caption{Averaged learning curves (with 95\% confidence intervals): effective training examples vs. mAP.}
\label{fig:learning_curves}
\end{figure}

\subsection{Setup}
\label{sec:exp:setup}

The learner agents start with relatively good, but still error-prone, priors of what bowls and stems and their attributes (e.g., ``short stem'') look like, but completely lacks the vocabulary, concepts and related visual features for the various glass types.
The prior knowledge is injected into the learner agents by exposing them to the full set of positive examples of stems and bowls in our data set, and randomly sampled non-instances for negative examples.
The average binary classification accuracies on balanced test sets were 98.11\% for the part concepts and 86.12\% for their attributes.

Our training and testing images are randomly generated from a simulation framework CoppeliaSim \cite{rohmer2013v}, using a toolkit for controlled sampling of 3D environments \cite{innes2021probrobscene}.
Each image features a scene of several objects from the restaurant domain laid on a tabletop (e.g., the image in Fig.~\ref{fig:arch}).
Each type of glass in our tabletop domain can be characterized by its parts having different attributes; see Appendix~\ref{app:glasses} for the complete list.
We implement simulated teachers in place of real human users for the experiments, which perform rule-based pattern matching just sufficient for participating as a teacher in our training dialogues.
Our implementation and datasets are publicly released in \url{https://github.com/itl-ed/ns-arch}.

The more distractor concepts we have, the more difficult the task becomes; difficulty scales roughly quadratically with respect to the number of concepts, since $C$ concepts enable ${C}\choose{2}$ different pairwise confusions.
Our experiments cover two levels of difficulty: \textsf{fineEasy} and \textsf{fineHard}.
For \textsf{fineEasy}, we set $(C,N_t,N_m)=(3,30,5)$, where target concepts are \{\texttt{brandy glass}, \texttt{burgundy glass}, \texttt{champagne coupe}\}.
For \textsf{fineHard}, we set $(C,N_t,N_m)=(5,60,10)$, where target concepts as those for \textsf{fineEasy} plus \{\texttt{bordeaux glass}, \texttt{martini glass}\}.

\subsection{Results and Discussion}

\begin{table*}
\small
\centering
\begin{tabular}{l|rrr|rrr}\toprule
    \multicolumn{1}{c|}{Task difficulty} & \multicolumn{3}{c|}{\textsf{fineEasy}} & \multicolumn{3}{c}{\textsf{fineHard}} \\ \midrule
    \multicolumn{1}{c|}{\# training examples} & 5 & 15 & 30 & 10 & 30 & 60 \\ \midrule
    \textsf{minHelp} & \qquad0.372 & \qquad0.478 & \qquad0.507 & \qquad0.253 & \qquad0.345 & \qquad0.355 \\
    \textsf{medHelp} & 0.371 & 0.494 & 0.529 & 0.241 & 0.346 & 0.426 \\
    \textsf{maxHelp\_semOnly} & 0.719 & 0.743 & 0.750 & 0.551 & 0.558 & 0.582 \\
    \textsf{maxHelp\_semNeg} & 0.727 & 0.797 & 0.805 & 0.572 & 0.636 & \textbf{0.681} \\
    \textsf{maxHelp\_semNegScal} & \textbf{0.744} & \textbf{0.803} & \textbf{0.811} & \textbf{0.574} & \textbf{0.649} & \textbf{0.681} \\ \bottomrule
\end{tabular}
\caption{Task performances of agents by mAP scores after different numbers of effective training examples.}
\label{tab:mAPs}
\end{table*}

Fig.~\ref{fig:learning_curves} and Tab.~\ref{tab:mAPs} display the averaged learning curves for the five strategy combinations in each task difficulty setting, along with 95\% confidence intervals.
It is obvious that learners exploiting the semantics of generic statements from \textsf{maxHelp} teachers are significantly faster in picking up new concepts, compared to the vision-only baseline configurations with \textsf{minHelp} or \textsf{medHelp} teachers.
Among the \textsf{maxHelp} results, the learners which extract and exploit additional, unstated information from the context (i.e., \textsf{semNeg} and \textsf{semNegScal}) outperform the learner \textsf{semOnly}, which doesn't exploit pragmatics.

\begin{figure}
\centering
\begin{subfigure}{0.44\textwidth}
    \includegraphics[width=\textwidth]{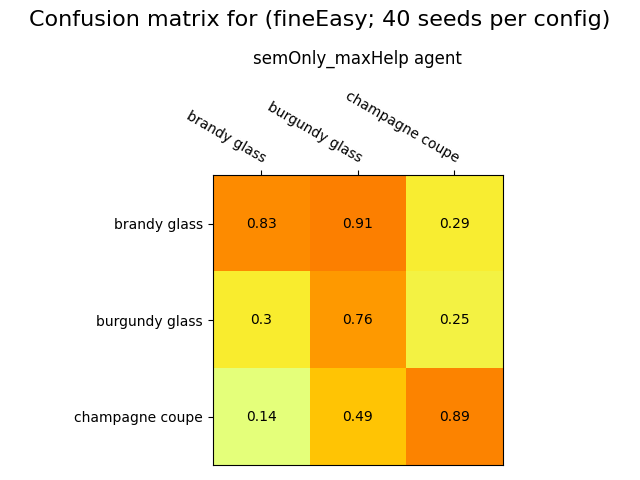}
    \caption{\textsf{maxHelp\_semOnly} on \textsf{fineEasy} difficulty.}
    \label{fig:confMats_semOnly}
\end{subfigure}
\par\bigskip
\begin{subfigure}{0.44\textwidth}
    \includegraphics[width=\textwidth]{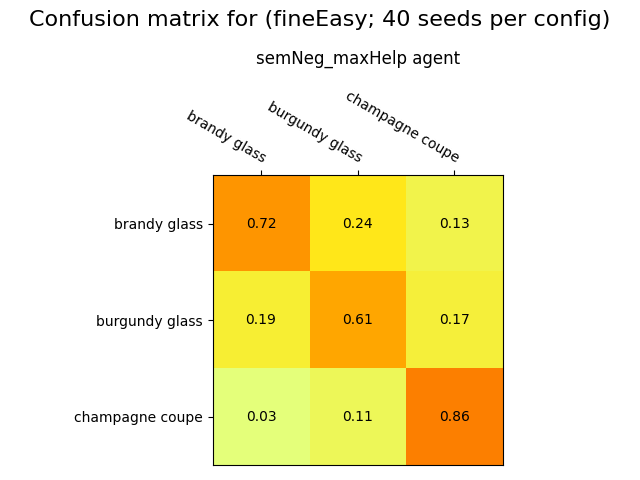}
    \caption{\textsf{maxHelp\_semNeg} on \textsf{fineEasy} difficulty.}
    \label{fig:confMats_semNeg}
\end{subfigure}
\caption{Averaged confusion matrices taken from the sequence-final evaluations for two configurations.}
\label{fig:confMats}
\end{figure}

Our error analysis reveals that the significant performance boosts enjoyed by \textsf{semNeg} and \textsf{semNegScal} learners comes from the ability to infer non-properties from property statements (i.e. $\psi^{neg}$ from $\psi$).
The confusion matrices reported in Fig.~\ref{fig:confMats} allow us to study the mechanism.
Specifically, notice how the \textsf{maxHelp\_semOnly} learner in Fig.~\ref{fig:confMats_semOnly} frequently misclassifies brandy glasses as burgundy glasses, whereas it is considerably less likely to make such mistakes in the opposite direction: 91\% vs. 30\%.
We can see this is because \textsf{semOnly} learners do not have access to the negative property of burgundy glasses of not having short stems ($\mathbb{G} O.burgundyGlass(O)\Rightarrow\neg haveShortStem(O)$).
Therefore, while \textsf{semOnly} learners can confidently dismiss instances of burgundy glasses as non-instances of brandy glasses, they are not able to dismiss instances of brandy glasses as non-instances of burgundy glasses.
We can observe in Fig.~\ref{fig:confMats_semNeg} that this is precisely remedied by \textsf{semNeg} and \textsf{semNegScal} learners, which are able to reliably distinguish the two types in both directions: 24\% vs. 19\%.

The difference between \textsf{semNeg} and \textsf{semNegScal} learner is more subtle.
Although their performances generally tend to converge after sufficient training, learners that exploit scalar implicatures seem to show higher data efficiency at earlier stages, especially in the \textsf{fineHard} task.
Nonetheless, the two learning curves have largely overlapping confidence intervals; we cannot make a strong scientific claim based on these results, and we will have to conduct experiments at a larger scale to corroborate this difference.

\section{Conclusion and Future Directions}
\label{sec:concl}

In this research, we have proposed an interactive symbol grounding framework for ITL, along with a neurosymbolic architecture for the learner agent.
We empirically showed that learners who can comprehend and exploit valid inferences from generic statements, including pragmatic content given their context of use, can learn to ground novel visual concepts more data-efficiently.
Our findings confirm it pays to study human-AI natural language interactions through the lens of discourse semantics, not only the truth conditions of isolated sentences but also their coherent connections to their context.

In future, we plan to relax some of many simplifying assumptions we made for controlled experiments, possibly exploring other domains.
For instance, the ideal assumption that teachers are infallible and communication is noise-free does not hold in most real-world scenarios \cite{appelgren2021symbol}.
Further, the set of linguistic constructions we have studied in this work is very constrained (as intended), and a natural next step is to accommodate a wider range of diverse and free-form NL constructions.
It is also a strong assumption that the learner agent already has relatively reliable beliefs about object part and concept attributes.
For example, if the learner does not know what the ``stem'' of a wine glass means, the absence of the concept must be resolved before communicating any generic characterizations involving stems.
Finally, our approach does not fully exploit the semantics of generic statements, which express qualitative rules that admit exceptions \cite{pelletier1997generics}.
The generic quantifier $\mathbb{G}$ did not play any significant role in this work.
% and we might as well have just used the universal quantifer $\forall$ instead to obtain the same results.
One major strength of ASP is that it is well suited for modeling non-monotonic inferences, and it would be interesting to study how to model ITL scenarios that can robustly address exceptions to generic rules.

\section*{Acknowledgements}

This work was supported by the UKRI Research Node on Trustworthy Autonomous Systems Governance and Regulation (grant EP/V026607/1) and by Informatics Global PhD Scholarships, funded by the School of Informatics at The University of Edinburgh.
We thank the anonymous reviewers for their feedback on an earlier draft of this paper, and Mattias Appelgren, Rimvydas Rubavicius and Gautier Dagan for continued feedback over the course of this research.

\bibliographystyle{acl_natbib}
\bibliography{anthology,acl2021}

\appendix

\section{Vision processing module: implementational details}
\label{app:vision}

\begin{figure*}
\centering
\includegraphics[width=0.9\textwidth]{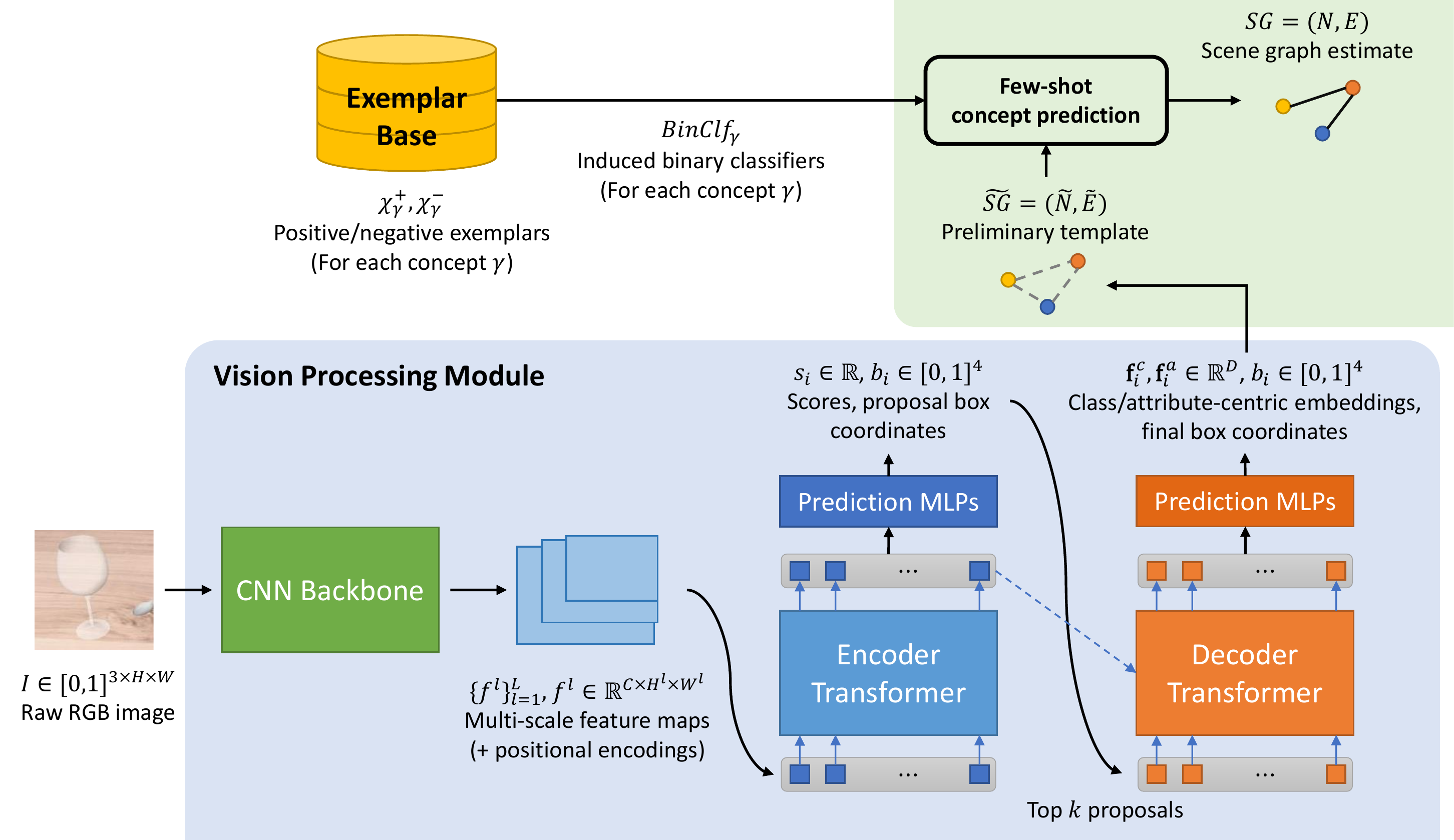}
\caption{A schematic of the structure of the vision processing module component in our agent architecture, and the pipeline through which raw visual inputs are processed into the final scene graph estimate.}
\label{fig:vision_module}
\end{figure*}

Our implementation of the few-shot neural vision processing module is based on the pretrained model of two-staged Deformable DETR \cite{zhu2021deformable}.
We train new lightweight multilayer perceptron (MLP) blocks for embedding image regions into low-dimensional feature spaces.
The MLP blocks replace the existing pretrained prediction heads that have fixed number of output categories, enabling metric-based few-shot predictions of incrementally learned visual concepts.

Let $C$, $A$ and $R$ denote open sets of visual concepts of different types: object classes\footnote{Intuitively corresponding to concepts denoted by nouns---e.g.,`brandy glass', `stem'.}, attributes\footnote{intuitively corresponding to concepts denoted by adjectives---e.g., `wide', `short'.} and pairwise relations\footnote{intuitively corresponding to concepts denoted by transitive verbs and adpositions---e.g., `have', `of'}.
In principle, we need one metric space for each concept type for their separate handling, hence three MLP blocks to train.
But for this work, $|R|=1$, where the only relation concept we need to capture is `have' (whole-part relationship).
We can make proxy predictions for the concept by the ratio of the area of bounding box intersection to the area of the candidate object part's bounding box.
Therefore, in the interest of simplicity, we prepare only two embedder blocks for $C$ and $A$ respectively; in future extensions where we need to deal with a truly open $R$, we will have to implement a relation-centric embedder block for $R$ as well.

Fig.~\ref{fig:vision_module} depicts how our vision module summarizes the raw RGB image input $\mathcal{I}\in[0,1]^{3\times H\times W}$ into a preliminary scene graph $\wtilde{SG}$, and then makes few-shot predictions to finally yield a scene graph $SG$.
$\mathcal{I}$ is first passed through the feature extractor backbone to produce $\{f_l\}_{l=1}^L$, a set of feature maps $f^l\in\mathbb{R}^{C\times H^l\times W^l}$ at $L$ different scales.
$\{f_l\}_{l=1}^L$ are flattened into a single sequence of input tokens (thus in $\mathbb{R}^{C\times \sum_l H^l\cdot W^l}$), combined with appropriate positional encodings and fed into the encoder.
We obtain from the encoder an objectness logit score $s_i$ and a proposal bounding box coordinate $\mathbf{b}_i\in [0,1]^4$ for the input tokens, out of which the top $k$ proposals with the highest $s_i$ scores are selected.
The selected proposals are fed into the decoder along with corresponding feature vectors to generate $\mathbf{f}_i^c, \mathbf{f}_i^a \in\mathbb{R}^D$, the class/attribute-centric embeddings of each input token, in addition to the (refined) bounding box coordinates $\mathbf{b}_i$.
The decoder outputs are collated into the preliminary scene graph template $\wtilde{SG}=(\tilde{N},\tilde{E})$.
$\tilde{N}$ is the node set containing $\mathbf{b}_i,\mathbf{f}_i^c,\mathbf{f}_i^a$ for each detected object.
$\tilde{E}$ is the edge set that \textit{would} contain pairwise relation-centric embeddings $\mathbf{f}_{i,j}^r$ for each pair of detected objects.
However, $\tilde{E}$ is essentially empty in our current implementation since as mentioned above, we fall back to proxy prediction by area ratio for the only relation concept of interest `have'.

For each visual concept $\gamma\in C,A(,R)$, the agent's visual XB stores $\chi_\gamma^{+/-}$, a set of positive/negative exemplars, which together naturally induce a binary classifier $BinClf_\gamma$.
We are free to choose any binary classification algorithm as long as it can return probability scores for concept membership from $\chi_\gamma^+$ and $\chi_\gamma^-$.
We use Platt-scaled SVM with RBF kernel \cite{platt1999probabilistic} in our implementation.
Then, $SG=(N,E)$ is generated from $\wtilde{SG}$ and a set of $BinClf_\gamma$'s, where $N$ and $E$ are each the scene graph node set and the scene graph edge set.
For each scene object, $N$ contains $\mathbf{c}_i\in[0,1]^{|C|}$ and $\mathbf{a}_i\in[0,1]^{|A|}$, each a vector designating the probabilistic beliefs of whether the object classifies as an instance of concepts in $C$ and $A$, as well as the box specification $\mathbf{b}_i$.
$E$ contains information about binary relationships between ordered pairs of objects, namely $\mathbf{r}_{i,j}\in[0,1]^{|R|}$, the probabilistic beliefs of whether the pair $(i,j)$ is an instance of concepts in $R$.
As a reminder, in our setting, $N$ is computed from $\tilde{N}$ and $BinClf_\gamma$ for each $\gamma\in C,A$, whereas $E$ is computed from bounding box area ratios.

Our new embedder blocks are trained on 50\% of the Visual Genome dataset \cite{krishna2017visual} with NCA loss objective \cite{goldberger2004neighbourhood} for metric learning, for 80,000 steps using SGD optimizer with the batch size of 64, the learning rate of $3\times 10^{-4}$ and the momentum factor of 0.1.
The prediction heads are then fine-tuned on our tabletop domain dataset\footnote{Excluding the fine-grained types of drinking glasses.} for 2,000 steps using Adam optimizer, with the batch size of 16, the initial learning rate of $2\times 10^{-4}$ and PyTorch default values\footnote{\url{https://pytorch.org/docs/stable/generated/torch.optim.Adam.html}} for the hyperparameters $\beta_1,\beta_2,\epsilon$.

\section{FOL representation of concept properties}
\label{app:fol_props}

In the main paper, we have represented the NL predication ``have short stems'' with an agglomerate predicate \textit{haveShortStem} for the sake of brevity, so that ``Brandy glasses have short stems'' would be translated into the \textsc{prop} $\mathbb{G} O.brandyGlass(O)\Rightarrow haveShortStem(O)$.
However, this is an oversimplification of what is actually happening under the hood in our implementation.
The predication ``have short stems'' ought to be broken down into its constituent meanings delivered by the individual tokens ``have'', ``short'' and ``stem'' respectively, for primarily two reasons: 1) they are the elementary units of concepts handled by the vision module and included in the output scene graphs, and 2) the object parts should be explicitly acknowledged as entities separate from the objects they belong to, and the generic \textsc{prop}s should model relations between objects and their parts (plus their attributes).

In light of this, we choose to read NL sentences of the form ``\{\texttt{object}\}s have \{\texttt{attribute}\} \{\texttt{part}\}s'' as follows: ``If $O$ is an \texttt{object}, there exists an entity $P$ such that $O$ has $P$ as its part, and $P$ is a \texttt{part} that is \texttt{attribute}''.
Then, for example, the sentence ``Brandy glasses have short stems'' would be represetned by the following \textsc{prop}:
\begin{align*}
&\mathbb{G} O.brandyGlass(O)\Rightarrow \\
&\qquad (\exists P.have(O,P),short(P),stem(P))
\end{align*}
or alternatively,
\begin{align*}
&\mathbb{G} O.brandyGlass(O)\Rightarrow \\
&\qquad have(O,f(O)),short(f(O)),stem(f(O))
\end{align*}
where $f$ is a skolem function that maps from the instance of brandy glass to its (only) short stem.
We opt for the latter option because it is more compliant with the formalism commonly used by logic programming methods, in which existential quantifiers are not admitted and variables are all implicitly universally quantified.

\section{More examples of grounding problems as probabilistic ASP programs}
\label{app:more_asp_exs}

Example~\ref{exm:2} below illustrates how lack of high-confidence observation of a short stem of a glass $o_1$ results in a weaker belief that $o_1$ is a brandy glass.

\begin{exm}
\label{exm:2}
The agent sees an object $o_1$ and initially estimates it's equally likely to be a brandy or burgundy glass.
The agent also notices with high confidence it does NOT have a short stem, and knows brandy glasses have short stems:
\begin{align}
\text{\emph{logit}}(0.61) &:\enspace brandyGlass(o_1). \\
\text{\emph{logit}}(0.62) &:\enspace burgundyGlass(o_1). \\
\text{\emph{logit}}(0.10) &:\enspace haveShortStem(o_1). \\
\text{\emph{logit}}(0.95) &:\enspace \bot\leftarrow brandyGlass(O), \notag\\
    &\qquad\qquad\text{\texttt{not} haveShortStem(O)}. \\
\text{\emph{logit}}(0.95) &:\enspace \bot\leftarrow haveShortStem(O), \notag\\
    &\qquad\qquad\text{\texttt{not} brandyGlass(O)}.
\end{align}
This results in $P_\Pi(brandyGlass(o_1))=0.20$, whereas $P_\Pi(burgundyGlass(o_1))=0.62$.
\end{exm}

Example~\ref{exm:3} shows how knowledge of \textit{negative} properties of an object class can affect symbolic reasoning.
The example supposes the agent's KB only consists of the knowledge ``Burgundy glasses do \textit{not} have short stems'', namely the \textsc{prop} $\mathbb{G} O.burgundyGlass(O)\Rightarrow\neg haveShortStem(O)$.
Note how we translate a generic \textsc{prop} whose $Cons$ is a negation of some $\mathcal{L}$-formula into probabilistic ASP rules.
Only rules penalizing deductive violations are generated, in which the negation ($\neg$) that wraps around $Cons$ `cancels out' the default negation \texttt{not}.
We do not generate rules for penalizing failures to explain $Cons$ from negative \textsc{prop}s, as abductive inferences of object classes from lack of properties would give rise to far-fetched conclusions: e.g., inferring something might be a banana because it does not have wheels.

\begin{exm}
\label{exm:3}
The agent sees an object $o_1$ and initially estimates it's equally likely to be a brandy or burgundy glass.
The agent also notices with high confidence it has a short stem, and knows burgundy glasses do NOT have short stems:
\begin{align}
\text{\emph{logit}}(0.61) &:\enspace brandyGlass(o_1). \\
\text{\emph{logit}}(0.62) &:\enspace burgundyGlass(o_1). \\
\text{\emph{logit}}(0.90) &:\enspace haveShortStem(o_1). \\
\text{\emph{logit}}(0.95) &:\enspace \bot\leftarrow burgundyGlass(O), \notag\\
    &\qquad\qquad\enspace haveShortStem(O).
\end{align}
This results in $P_\Pi(brandyGlass(o_1))=0.61$, whereas $P_\Pi(burgundyGlass(o_1))=0.19$.
\end{exm}

Note that knowledge about brandy glasses do not affect the likelihood of $o_1$ being a burgundy glass, and \textit{vice versa}: i.e., for an object, the events of being a brandy glass vs. a burgundy glass are independent.
This is because the KBs in the examples do not introduce any type of dependency between the two glass types.
For instance, if we inject mutual exclusivity relation between the two types in the KB, both probability values $P_\Pi(brandyGlass(o_1))$ and $P_\Pi(burgundyGlass(o_1))$ would be affected by knowledge about either.

\section{Task domain: Fine-grained types of drinking glasses to distinguish}
\label{app:glasses}

\begin{table}[ht]
\small
\centering
\begin{adjustbox}{width=0.48\textwidth}
\begin{tabular}{p{0.7in}|p{1.5in}|M{0.8in}}
\toprule
\multicolumn{1}{c|}{Type} & \multicolumn{1}{c|}{Properties} & \multicolumn{1}{c}{Sample image} \\
\midrule
\texttt{bordeaux glass} & Bowl: elliptical, tapered. & \includegraphics[height=0.6in]{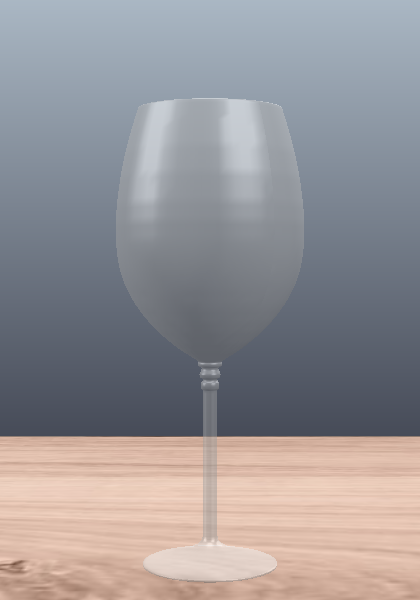} \\ 
\texttt{brandy glass} & Bowl: wide, tapered, round. \newline Stem: short. & \includegraphics[height=0.6in]{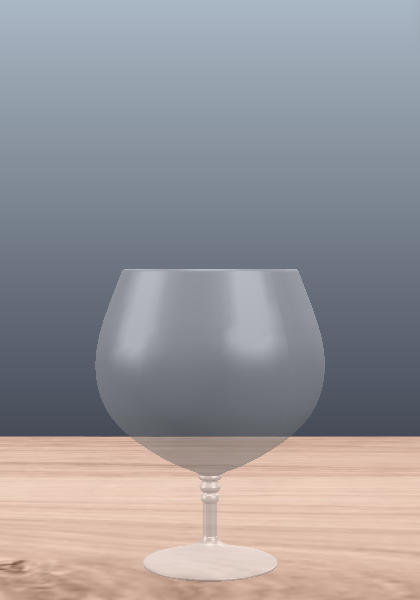} \\
\texttt{burgundy glass} & Bowl: wide, tapered, round. & \includegraphics[height=0.6in]{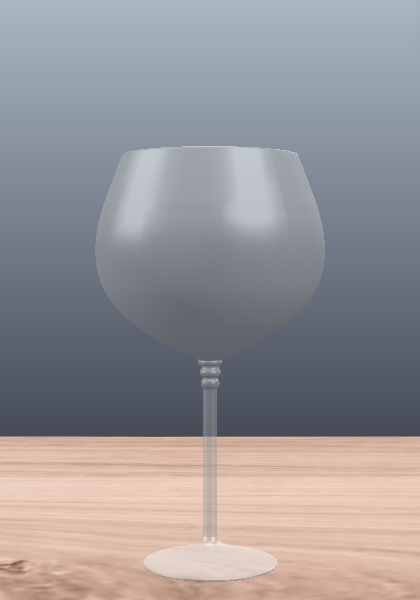} \\
\texttt{champagne coupe} & Bowl: broad, round. & \includegraphics[height=0.6in]{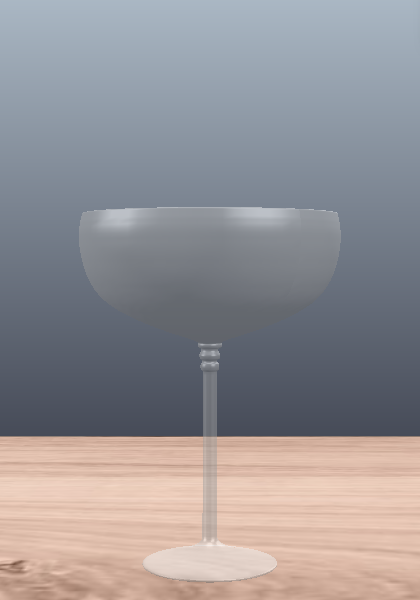} \\
\texttt{martini glass} & Bowl: broad, conic. & \includegraphics[height=0.6in]{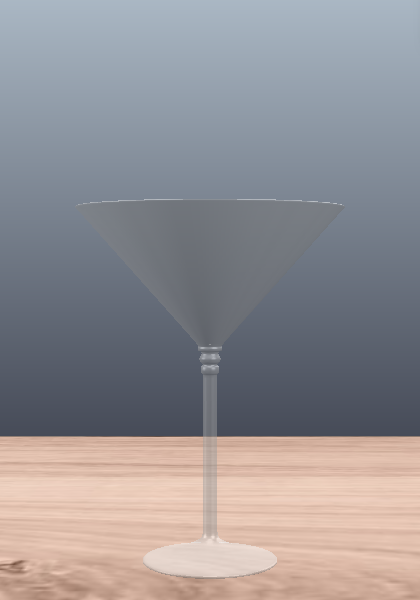} \\
\bottomrule
\end{tabular}
\end{adjustbox}
\caption{Fine-grained types of drinking glasses modeled in our tabletop domain. (Note only brandy glasses have characteristic stems, whereas bowls of all glass types can be characterized by some set of attributes.)}
\label{tab:glasses}
\end{table}

\begin{table*}
\small
\centering
\begin{adjustbox}{width=\textwidth}
\begin{tabular}{l|l|l}\toprule
    \multicolumn{1}{c|}{Confusion} & \multicolumn{1}{c|}{\texttt{champagne coupe - burgundy glass}} & \multicolumn{1}{c}{\texttt{burgundy glass - bordeaux glass}} \\ \midrule
    \multirow{9}{*}{KB state} & $\mathbb{G} O.champagneCoupe(O)\Rightarrow haveBroadBowl(O)$ & $\mathbb{G} O.burgundyGlass(O)\Rightarrow$ \\
    & $\mathbb{G} O.burgundyGlass(O)\Rightarrow$ & \qquad\qquad$haveWideBowl(O),haveRoundBowl(O)$ \\
    & \qquad\qquad$haveWideBowl(O),haveTaperedBowl(O)$ & $\mathbb{G} O.bordeauxGlass(O)\Rightarrow haveEllipticalBowl(O)$ \\
    & $\color{red} \mathbb{G} O.burgundyGlass(O)\Rightarrow\neg haveBroadBowl(O)$ & $\color{red} \mathbb{G} O.bordeauxGlass(O)\Rightarrow$ \\
    & $\color{red} \mathbb{G} O.champagneCoupe(O)\Rightarrow$ & \qquad\quad$\color{red} \neg(haveWideBowl(O),haveRoundBowl(O))$ \\
    & \qquad\quad$\color{red} \neg(haveWideBowl(O),haveTaperedBowl(O))$ & $\color{red} \mathbb{G} O.burgundyGlass(O)\Rightarrow\neg haveEllipticalBowl(O)$ \\
    & & $\color{blue} \underline{\mathbb{G} O.bordeauxGlass(O)\Rightarrow}$ \\
    & & \qquad\qquad$\color{blue} \underline{haveWideBowl(O),haveTaperedBowl(O)}$ \\
    & & $\color{blue} \mathbb{G} O.bordeauxGlass(O)\Rightarrow\neg haveBroadBowl(O)$ \\ \bottomrule
\end{tabular}
\end{adjustbox}
\caption{An example illustration of how \textsf{semNegScal} learners can infer incorrect and unintended knowledge. The underlined \textsc{prop} denotes a generic rule which is neither correct nor intended by the teacher.}
\label{tab:impl_scl_failure}
\end{table*}

Tab.~\ref{tab:glasses} lists the set of fine-grained types of drinking glasses that are modeled in our simulated tabletop domain, along with their properties and sample images.
3D meshes of the glasses are obtained from a website that lists stock models made by third-party providers,\footnote{\url{https://www.turbosquid.com/3d-models/wine-glasses-3d-1385831}} then imported into the simulation environment.

As illustrated, properties of each fine-grained type comprise its part attributes.
For instance, the full set of properties of a brandy glass could be expressed as a set \{(\text{wide}, \text{bowl}), (\text{tapered}, \text{bowl}), (\text{round}, \text{bowl}), (\text{short}, \text{stem})\}.
When asked, our simulated teacher computes the answer to ?\texttt{conceptDiff} \textsc{ques} as pairwise symmetric differences between property sets: e.g., for ?\texttt{conceptDiff}(\textit{brandyGlass},\textit{burgundyGlass}), we obtain \{(\text{short}, \text{stem})\} for brandy glasses and $\varnothing$ for burgundy glasses.

These properties of glasses did not ship with the 3D models; instead, we hand-coded them based on information available on the internet.
We have put effort to prepare an annotation scheme that is faithful to properties of the glasses in the reality, yet the domain knowledge may still have inconsistency against the `ground-truth'---any error in that regard remains our own.

\section{Rule acquisition by inference of implicatures and failure case analysis}
\label{app:impl_fail}

In this work, we assume that all generic NL statements given by the teacher are characterizations of object classes by their positive properties (those described in Appendix~\ref{app:glasses}), and statements of \textit{negative} properties are never explicitly provided.
This reflects the fact that we usually characterize things by their positive properties rather than by their negative properties because the former generally have more determining power \cite{zangwill2011negative}.
Therefore, in our experiments, negative properties can be obtained only by virtue of inference of implicatures.
That is, for example, only \textsf{semNeg} or \textsf{semNegScal} learners have access to the negative \textsc{prop} $\mathbb{G} O.burgundyGlass(O)\Rightarrow\neg haveShortStem(O)$.

Nevertheless, \textsf{semNegScal} learners risk acquisition of incorrect and unintended knowledge when they make inferences of scalar implicatures.
To see this, study the example illustrated in Tab.~\ref{tab:impl_scl_failure}, where two successive confusions take place in the order of \texttt{brandy glass} vs. \texttt{burgundy glass} and then \texttt{burgundy glass} vs. \texttt{champagne coupe}.
In the example, the underlined \textsc{prop} successfully infiltrates into the learner's KB without being suppressed by explicitly stated \textsc{prop}s or their negative implicature counterparts.
This is why inference of the scalar implicatures should be cancellable, so that they can be retracted in the face of contradictory evidence.
In our implementation, this is achieved by periodically inspecting the KB entries against the episodic memory, removing any rules whose counterexamples are found.

\end{document}